\documentclass{article}

\usepackage[preprint]{neurips_2026}

\usepackage[utf8]{inputenc} 
\usepackage[T1]{fontenc}    
\usepackage{hyperref}       
\usepackage{url}            
\usepackage{booktabs}       
\usepackage{makecell}     
\usepackage{pifont}       
\newcommand{\cmark}{\ding{51}}
\newcommand{\xmark}{\ding{55}}
\usepackage{amsmath}        
\usepackage{amsfonts}       
\usepackage{nicefrac}       
\usepackage{microtype}      
\usepackage{xcolor}         
\usepackage{graphicx}
\usepackage{comment}
\usepackage{caption}

\usepackage{xparse}
\RenewDocumentCommand{\paragraph}{s o m}{%
  \par\noindent\textbf{#3}\ %
}

\title{Fast 4D Mesh Generation by Spatio-Temporal Attention Chains}

\author{%
  Dvir Samuel$^{1}$ \quad
  Yuval Atzmon$^{1}$ \quad
  Gal Chechik$^{1,2}$ \quad
  Yoni Kasten$^{1}$ \\
  $^{1}$NVIDIA Research, Tel-Aviv, Israel \\
  $^{2}$Bar-Ilan University, Ramat-Gan, Israel
}

\begin{document}

\maketitle

\begin{abstract}
4D mesh generation has recently emerged as a powerful paradigm for recovering dynamic 3D structure from videos, but existing methods remain slow, computationally expensive, and difficult to scale to longer sequences. We introduce a training-free approach that accelerates 4D mesh generation while improving temporal correspondence quality. Our key observation is that temporal correspondences emerge inside a 4D backbone long before its generated meshes become visually accurate. We exploit this with a general framework we call Spatio-Temporal Attention Chain which propagates information across space and time. Starting from vertices on an anchor mesh, the chain maps vertices to latent tokens. It then follows temporal correspondences in latent space, and recovers frame-specific vertices through latent-to-vertex attention. This design avoids expensive explicit matching while preserving anchor mesh details and thereby improving dynamic mesh geometry and temporal consistency.

Compared to state-of-the-art, our method generates a 4D mesh in 9 seconds, achieving a $13\times$ speedup while producing higher-quality results. Moreover, our approach scales to videos up to $16\times$ longer without degrading mesh quality. Beyond generation, the improved correspondences enable competitive zero-shot performance on two downstream tasks: 2D object tracking and 4D tracking. We further show that our framework enables reliable camera estimation, a capability not supported by prior 4D mesh generation methods. \href{https://research.nvidia.com/labs/par/fast4dmesh/}{\textcolor{blue}{Project Page}}
  \end{abstract}

\section{Introduction}
\label{sec:intro}

Understanding a dynamic 3D world is a central goal of computer vision, and a foundation for embodied AI, physical reasoning, simulation, and virtual reality. Yet progress on dynamic 3D lags far behind images and videos. The bottleneck is data scarcity: high-quality 4D data must capture both 3D structure and motion over time, making it rare and expensive to acquire. This motivates recovering 4D from ordinary videos, a far more scalable source of motion and shape~\citep{consistent4d,dreamgaussian4d,4dgen,sv4d,cat4d}. In this work, we focus on \emph{video-to-dynamic-mesh reconstruction}: recovering a temporally coherent 3D mesh sequence from a video of a moving object.

Reconstructing dynamic geometry from monocular video is challenging because the model must infer detailed 3D geometry in every frame while preserving surface identity across time, a requirement that is hard to learn under scarce 4D supervision~\citep{v2m4,dreammesh4d,lim,shapegen4d,actionmesh}. ActionMesh~\citep{actionmesh} resolves this with a staged design. A 4D generative diffusion backbone first lifts the video into 3D latents with frame-specific topology; a separate network then animates an anchor mesh to enforce shared connectivity. While effective, this design is costly. The generative stage requires significant time to produce high-quality geometry; the second stage adds non-end-to-end training; and the full pipeline remains tied to scarce 4D supervision. Beyond these costs, the output mesh lies in an arbitrary coordinate frame with no link back to input pixels, preventing downstream applications like 4D and 2D tracking, camera recovery, or scene composition~\citep{tapir,cotracker,spatracker,tracksto4d,vggt}. Finally, training on short clips makes this drift or collapse on longer videos.

That additional stage nevertheless reveals an important fact: the geometry comes from a simpler image-to-3D anchor, while the heavy 4D generative backbone contributes only a 4D motion prior. This raises a natural question: how can we apply that prior directly to the anchor, without a separate neural animator?

We show that the answer lies inside the 4D generative backbone itself: our key observation is that useful temporal correspondences already emerge when we run denoising with as few as four denoising steps, where attention patterns already link anchor-frame 3D tokens to matching tokens in later frames. We expose this signal through a \emph{Spatio-Temporal Attention Chain}. At a high level, the chain treats \textit{attention as a soft Markov transport}: each attention row is a probability distribution over latent tokens, so multiplying attention maps gives the probability of moving from one representation to the next. Concretely, it links an anchor-frame vertex to an anchor-frame latent token ($V_a \to Z_a$), transports it across time to frame $f$ via temporal self-attention ($Z_a \to Z_f$), and projects it back to a 3D point in the target frame ($Z_f \to V_f$).

This shifts the role of the backbone: instead of producing a high-resolution mesh per frame with 30 denoising steps, we run the denoiser with four steps and read the correspondence field it already computes. By tracking sparse landmarks through the chain and lifting them to the anchor mesh via geodesic-rigid skinning, we drop the second stage entirely. Generation time therefore falls from nearly two minutes to roughly nine seconds, preserving topology by construction and improving 3D accuracy.
Beyond speed, the chain naturally connects 2D patches, latent tokens, and mesh vertices, unlocking three additional capabilities. First, when extending generation from $16$ to $240$ frames autoregressively, reinforcing the strongest correspondences during denoising substantially reduces geometric drift without retraining. Second, different chain compositions yield competitive zero-shot 3D point trajectories (4D tracking) and 2D point trajectories. Third, the resulting 2D-to-3D matches allow recovering per-frame cameras, placing the mesh back into a reconstructed scene (Fig.~\ref{fig:mesh_placement}).

\noindent\textbf{Contributions.}
(1) We identify spatio-temporal attention chains as a hidden correspondence signal linking pixels, latent tokens, and mesh vertices inside a 4D generative backbone.
(2) We propose a training-free framework for 4D generative backbones that tracks sparse landmarks through these chains and lifts them to a full animated mesh, cutting inference from 120s to 9s at favorable quality.
(3) We improve autoregressive generation by reinforcing these correspondences to reduce drift, enabling coherent $16\times$ longer sequences.
(4) We show that the same chains yield several capabilities missing in prior work: competitive zero-shot 4D and 2D point tracking, as well as camera recovery from 2D-3D matches.

\section{Related Work}

\paragraph{Image-to-3D generative backbones.} Our chain reads attention weights off a VecSet-style 3D decoder~\citep{shape2vecset}, which reconstructs geometry by cross-attending 3D query points to a compact set of latent tokens encoding the shape. We instantiate on TripoSG~\citep{triposg}, a flow-based image-to-3D generator producing high-fidelity meshes from a single image. The same decoder structure underlies CLAY~\citep{clay}, Craftsman~\citep{craftsman}, Dora-VAE~\citep{doravae}, and Hunyuan3D~\citep{hunyuan3d}, and our chain could be adapted to other generators sharing this structure. Other representation classes include Trellis's~\citep{trellis} sparse structured latents (SLAT) based on active voxels, LRM's~\citep{lrm} triplanes, LGM's~\citep{lgm} Gaussian primitives, and AssetGen's~\citep{assetgen} PBR-textured meshes.

\paragraph{Video-to-4D generation.}
\label{sec:rw_video4d}
Per-scene optimization pipelines~\citep{consistent4d,dreamgaussian4d,4dgen,sc4d,stag4d,vidu4d,4diffusion,diffusion4d} distill dynamic 3D from video using diffusion priors, taking minutes to hours per clip. Multi-view video diffusion~\citep{sv4d,sv4d2,cat4d,animate3d} generates feed-forward novel-view sequences but still needs per-scene optimization for 4D. Feed-forward 4D methods predict spatial primitives directly, typically in topology-free spaces: L4GM~\citep{l4gm} and 4DGT~\citep{four_dgt} produce Gaussian sequences, Motion2VecSets~\citep{motion2vecsets} denoises vector sets, and ShapeGen4D~\citep{shapegen4d} adds temporal attention to a 3D generator; all decode frames independently, lacking shared topology.
Prior methods therefore add an explicit topology stage: ActionMesh~\citep{actionmesh} learns a temporal 3D autoencoder that deforms a reference mesh via per-frame anchor displacements, $\mathbf{V}_f = \mathbf{V}_a + \Delta_f$, while optimization-based methods~\citep{v2m4,dreammesh4d,lim} impose topology or temporal consistency through registration, deformation, or optimized implicit representations.
In contrast, our approach extracts dense correspondences directly from the temporal backbone, bypassing any separate topology-enforcing or animation stage. A related line animates a given mesh via predicted skeletons~\citep{makeitanimatable,magicarticulate,riganything,riggs} or deformation fields~\citep{smf,driveanymesh,animateanymesh}; these methods assume clean input assets and predict explicit skeletons and skinning weights.

\paragraph{Emergent Correspondences in Diffusion Features:}
Two recent lines tap frozen diffusion models for zero-shot correspondence.
One matches features as descriptors: DIFT~\citep{dift} on UNet activations, Diff3F~\citep{dutt2024diff3f} lifting them onto 3D shapes, MbQ \citep{motionbyqueries} on video-DiT queries for Q-injection, and Track4Gen~\citep{track4gen} via an auxiliary tracking loss.
The other reads attention weights directly: CAMEO~\citep{cameo} in multi-view 3D attention, DiTFlow~\citep{ditflow} as a per-clip optimization loss, and DiffTrack \citep{difftrack} on video-DiT temporal-matching layers; Point Prompting~\citep{pointprompting} sidesteps both via counterfactual prompting.
We instead \textit{compose} three attention maps of a frozen 4D generator -- vertex-to-token, temporal token-to-token, and token-to-surface -- into a $V_a\!\to\!Z_a\!\to\!Z_f\!\to\!V_f$ chain yielding correspondences tied to an anchor mesh's surface through a forward pass -- no optimization, no external tracker.

\paragraph{Attention Control in Diffusion Models:}
A complementary line manipulates or analyzes frozen attention. Several methods reweight cross/self-attention~\citep{hertz2022prompt,samuel2025omnimattezero} for editing, inject self-attention features~\citep{tumanyan2023plug} for structure, or share self-attention across images~\citep{masactrl,consistory} for identity; TiARA~\citep{tiara} suppresses temporal attention weights for extended video generation. A related thread treats attention rows as probability distributions and composes them to trace information flow within a single transformer~\citep{abnar2020quantifying,chefer2021generic,erel2025attention}. Building on this view, we compose attention across separately-trained modules and modalities of a 4D pipeline -- vertex-to-token, token-to-token, token-to-surface -- and reinforce reliable matches to stabilize long sequences.

\paragraph{Point tracking and monocular 4D geometry.} Our method outputs 2D and 3D point trajectories. Supervised 2D trackers~\citep{pips,tapir,bootstap,cotracker,cotracker3,locotrack,dot,alltracker} are driven by standard benchmarks~\citep{tapvid,pointodyssey}; 3D trackers~\citep{spatracker,spatrackerv2,tapip3d} update point clouds, while 4RC~\citep{4rc}, Trace-Anything~\citep{traceanything}, and TracksTo4D~\citep{tracksto4d} predict motion fields and MegaSaM~\citep{megasam} runs deep visual SLAM. A separate line predicts metric pointmaps, introduced by DUSt3R~\citep{dust3r} and extended for dynamic scenes~\citep{stereo4d,st4rtrack,monst3r,cut3r,geometrycrafter}; closest in spirit, Easi3R~\citep{easi3r} achieves 4D reconstruction via training-free attention adaptation of DUSt3R. These methods either require per-frame mesh reconstruction from scattered points or depend on pointmap supervision. In contrast, our approach requires no tracker or pointmap supervision. Furthermore, our single forward pass directly outputs the skinned mesh alongside the 2D-3D matches needed for PnP+RANSAC~\citep{lepetit2009epnp,fischler1981ransac} camera pose estimation.

\section{4D Mesh Generation: Preliminaries and Notations}
\label{sec:preliminaries}
Video-to-dynamic-mesh methods map a video of $F$ frames, to a temporally coherent 4D mesh sequence $\mathcal{M}_{1:F}=\{(\mathbf{V}_f,\mathcal{F})\}_{f=1}^{F}$, where $\mathcal{F}$ and $\mathbf{V}_f\in\mathbb{R}^{|V|\times 3}$ define a fixed topology and shared vertex identities across time.

We denote attention between a query sequence $\mathbf{X}$ and context sequence $\mathbf{Y}$ by:
\begin{equation}
    A_{\mathbf{X}\rightarrow\mathbf{Y}} =
    \textit{softmax}((\mathbf{X}\mathbf{W}_Q)(\mathbf{Y}\mathbf{W}_K)^\top / \sqrt{d_k}),
    \quad
    \operatorname{Attn}(\mathbf{X}, \mathbf{Y}) =
    A_{\mathbf{X}\rightarrow\mathbf{Y}} \cdot (\mathbf{Y}\mathbf{W}_V)
\end{equation}
    where $\mathbf{W}_Q$, $\mathbf{W}_K$, and $\mathbf{W}_V$ project inputs to queries, keys, and values.

Recent pipelines~\citep{actionmesh,v2m4,mesh4d} employ a three-staged approach: (0) an image-to-3D model reconstructs initial reference geometry, (I) temporal or independent generators produce per-frame 3D representations, and (II) a final topology-preserving stage aligns the vertices through time so all frames share the same connectivity.

    \paragraph{Stage 0: Image-to-3D anchor reconstruction.}
    An image-to-3D denoiser model (e.g.~\citep{triposg}) reconstructs an anchor mesh $\mathcal{M}_a=(\mathbf{V}_a,\mathcal{F})$ with shape latent $\mathbf{z}_a\in\mathbb{R}^{N\times d}$ composed of $N$ tokens of dimension $d$.
    Then, a VAE's transformer decoder expresses each anchor vertex as an attention-weighted combination of latent tokens, yielding $A_{V_a\rightarrow Z_a} \in \mathbb{R}^{|V_a| \times N}$. Image cross-attention similarly links anchor image patches $P_a$ to the same tokens, giving $A_{P_a\rightarrow Z_a}$.

    \paragraph{Stage I: Video-to-4D mesh generation.} Given the anchor latent $\mathbf{z}_a$ and the input video, a temporal denoiser $\Phi_\theta$ predicts one latent $\mathbf{z}_f$ for each frame $f$. Inflated self-attention inside $\Phi_\theta$ links tokens in $\mathbf{z}_a$ to tokens in $\mathbf{z}_f$, giving $A_{Z_a\rightarrow Z_f} \in \mathbb{R}^{N \times N}$.

    \paragraph{Stage II: Topology-consistent decoding.}
To maintain consistent topology, prior pipelines add a topology-preserving stage to predict per-frame displacements using learned decoders~\citep{mesh4d,actionmesh} or test-time optimization~\citep{v2m4,dreammesh4d}. In contrast, we drop this stage, recovering anchor motion directly from attention-chain correspondences.

\begin{figure}[t!]
    \centering
    \includegraphics[width=1\linewidth]{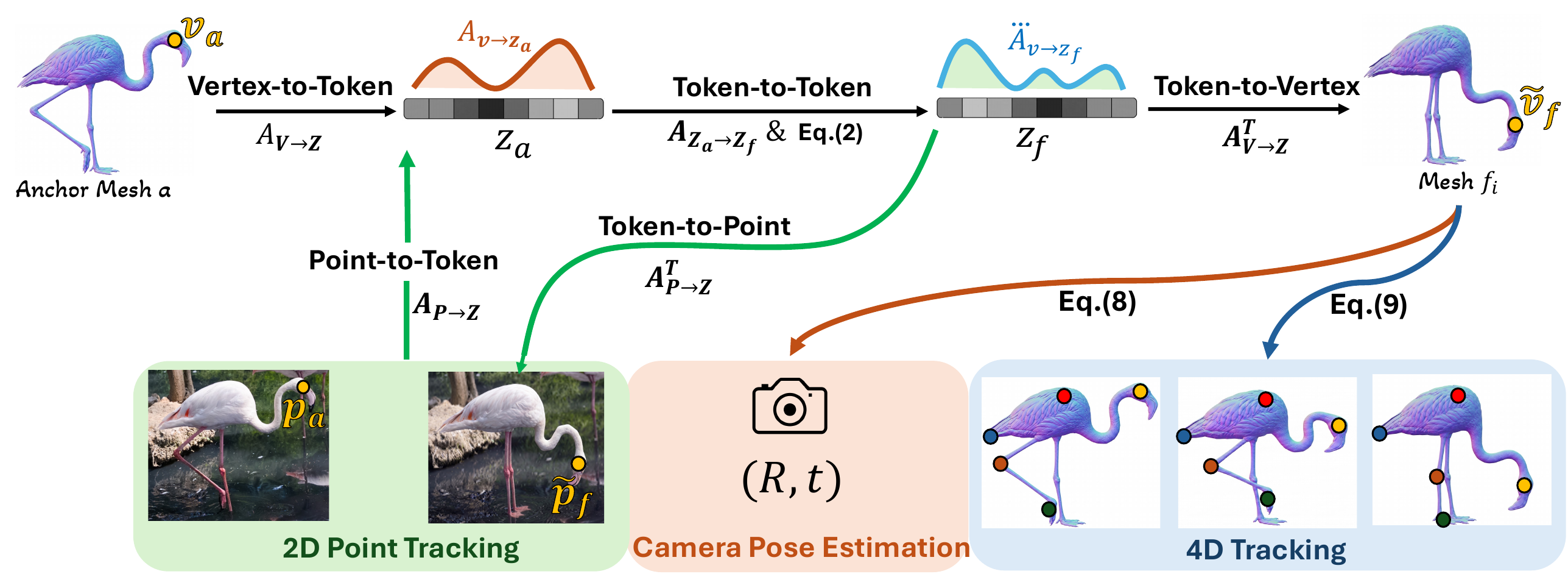}
   \caption{\textbf{Method overview.}
Our attention chain follows a point through the frozen 4D generator: from an anchor mesh vertex to latent tokens, across time to target-frame tokens, and back to a target mesh vertex. Image-patch endpoints give analogous chains for 2D tracking, camera pose estimation, and 4D tracking, without additional training.}
    \label{fig:method}
\end{figure}

\section{Method}
\label{sec:method}
Current staged pipelines~\citep{actionmesh,v2m4} treat geometry generation and animation as separate tasks. However, relying on a dedicated Stage II requires an entirely separate network, adding significant computational overhead during both training and inference. Moreover, these pipelines typically remain restricted to short, drift-prone temporal windows (Fig.~\ref{fig:long_video}). We aim to accelerate and scale topology-preserving 4D generation without any additional training. Our core observation is that a frozen pipeline like ActionMesh~\citep{actionmesh} inherently encodes temporal tracking within its features. Instead of relying on a learned decoder, we extract 3D correspondences directly during the denoising process (stages 0 and I) via an \emph{attention chain}. Conceptually, each attention matrix is a soft transition map, and multiplying them transports probability mass from anchor vertices, through latent tokens, to target surface points. This chain maps anchor vertices ($V_a$) to latent tokens ($Z_a$), transports them across frames via temporal self-attention ($Z_a\rightarrow Z_f$), and projects them back to target surface points $V_f$ at frame $f$:\begin{equation*}
    V_a \rightarrow Z_a \rightarrow Z_f \rightarrow V_f\end{equation*}These correspondences emerge within just a few denoising steps. We then animate the anchor mesh using a fast closed-form deformation model. By strictly reusing the constant face set $\mathcal{F}$, we guarantee perfect topology consistency by construction.

\subsection{Correspondence from the Attention Chain}
\label{sec:attention_chain}

For each anchor vertex $v$ and target frame $f$, we seek a target surface point $\tilde{v}_f$. Instead of training a separate deformation network, we
establish tracking with an attention chain (Fig.~\ref{fig:method}).
The chain links the anchor and target geometries through intermediate representations by sequentially multiplying the backbone's internal spatial and temporal attention maps. This composes localized attention steps into a dense correspondence map. We assemble the chain from three components.

\textbf{(1) Vertex-to-token attention ($V_a \rightarrow Z_a$).} During Stage 0 (Sec.~\ref{sec:preliminaries}), the 3D decoder yields the cross-attention matrix $A_{V_a\rightarrow Z_a} \in \mathbb{R}^{|V_a| \times N}$. Since the softmax normalization is applied over the latent key dimension, each row $A_{V_a\rightarrow Z_a}[v, :]$ forms a valid probability distribution. This row explicitly describes which latent tokens explain anchor vertex $v$, where each entry represents the probability that anchor token $t$ relates to that specific vertex.

\textbf{(2) Token-to-token temporal attention ($Z_a \rightarrow Z_f$).} During the denoising step of Stage I, the inflated temporal self-attention layers process all frames simultaneously. For a given target frame $f$, we extract the attention weights linking anchor-frame tokens to frame-$f$ tokens to yield $A_{Z_a\rightarrow Z_f} \in \mathbb{R}^{N \times N}$. This matrix governs the transfer of structural information from the anchor frame to the target frame at the latent token level.

\textbf{(3) Token-to-surface attention ($Z_f \rightarrow V_f$).} For target frame $f$, the 3D decoder turns $Z_f$ into an implicit field. We extract candidate surface points $S_f=\{x_u^{(f)}\}_{u=1}^{|V_f|}$ from this field and query them against the frame-$f$ latent tokens. The resulting cross-attention matrix $A_{Z_f\rightarrow V_f}^T \in \mathbb{R}^{|V_f| \times N}$ relates each candidate surface point $u$ to these tokens.

\textbf{Composing the attention chain.} We compose the attention matrices above to map anchor vertex $v$ to frame $f$. The row $A_{V_a\rightarrow Z_a}[v,:]$ gives weights over anchor tokens. Multiplying this row by $A_{Z_a\rightarrow Z_f}$ transfers those weights to frame-$f$ tokens $t' \in Z_{f}$:
\begin{equation}
   \dddot{A}_{v,Z_f}(t') = \sum_{t=1}^{N} A_{V_a\rightarrow Z_a}[v,t]\, A_{Z_a\rightarrow Z_f}[t,t'],
   \end{equation}
where $t$ indexes anchor-frame tokens and $t'$ indexes tokens in frame $f$. The vector $\dddot{A}_{v,Z_f}$ is therefore a probability distribution over frame-$f$ tokens. A candidate surface point $x_u^{(f)} \in S_f$ is likely to match $v$ when its token-level attention agrees with $\dddot{A}_{v,Z_f}$, so we score it by their inner product:
\begin{equation}
s_{v,f}(u) = \sum_{t'=1}^{N}  \dddot{A}_{v,Z_f}(t')\, A_{Z_f\rightarrow V_f}^T [u,t'].
\end{equation}
Finally, we obtain the correspondence $\tilde{v}_f$ as a sharp softmax blend over the top-scoring surface points:
\begin{equation}
\tilde{v}_f = \sum_{u\in\mathcal{N}_{v,f}} \pi_{v,f}(u)\,x_u^{(f)}, \qquad \pi_{v,f}(u) = \frac{\exp(s_{v,f}(u)/\tau)}{\sum_{q\in\mathcal{N}_{v,f}}\exp(s_{v,f}(q)/\tau)}.
\end{equation}
Here $\mathcal{N}_{v,f}$ denotes the localized subset of top-scoring surface points and $\tau$ is a temperature hyperparameter. We also define a confidence score $c_v^{(f)} = \max_{u} s_{v,f}(u)$ to be used later for mesh animation.

This construction has two key properties. First, both endpoint attentions come from the same 3D decoder, so anchor vertices and target surface points are compared in a shared token--geometry space. Second, $\tilde{v}_f$ is computed only from the top-scoring surface samples $\mathcal{N}_{v,f}$, keeping the correspondence on the target surface and reducing drift to unrelated regions.

The next step lifts these sparse correspondences to a full animated mesh while preserving the anchor topology efficiently and without any additional model training.

\subsection{Topology-Preserving Animation}\label{sec:animation}
In early experiments, we observed that directly querying all anchor vertices and simply mapping them to their target positions using our dense correspondences produced noisy results. Instead, we obtain topology-preserving animation by tracking a sparse set of control landmarks and lifting their motion to the full mesh in three steps:

\textbf{1. Landmark Extraction and Filtering:} We sample a sparse set of $K$ control landmarks on the anchor mesh by farthest point sampling. We extract their trajectories across frames via the attention chain, assigning confidence scores and rejecting physically implausible displacements as outliers.

\textbf{2. Temporal Smoothing:} To ensure fluid motion, we apply a confidence-weighted 1D Gaussian temporal smoothing to each landmark's trajectory independently. This bridges gaps caused by outlier removal by interpolating each landmark from nearby reliable frames.

\textbf{3. Mesh Deformation:} Finally, we propagate the smoothed landmark motions to the dense mesh using Geodesic Rigid Skinning~\cite{sumner2007embedded}. For each vertex, we compute a local rigid transformation (rotation and translation)
from its closest landmarks under geodesic distance, which is measured along the mesh surface. This prevents motion from leaking between spatially close but disconnected parts, such as an arm and torso, while the local-rigid transform preserves volume and avoids the shrinkage artifacts often caused by linear blend skinning.
This pipeline yields a temporally coherent animated mesh $\hat{\mathcal{M}}_f = (\hat{\mathbf{V}}_f, \mathcal{F}_a)$ that strictly maintains the anchor topology. Full details of temporal smoothing and the weighted Procrustes skinning formulation are deferred to Appendix~\ref{sec:appendix_animation}.

\subsection{Scaling to Longer Sequences}
\label{sec:scaling_video}

\begin{figure}[t!]
    \centering
    \includegraphics[width=1\linewidth]{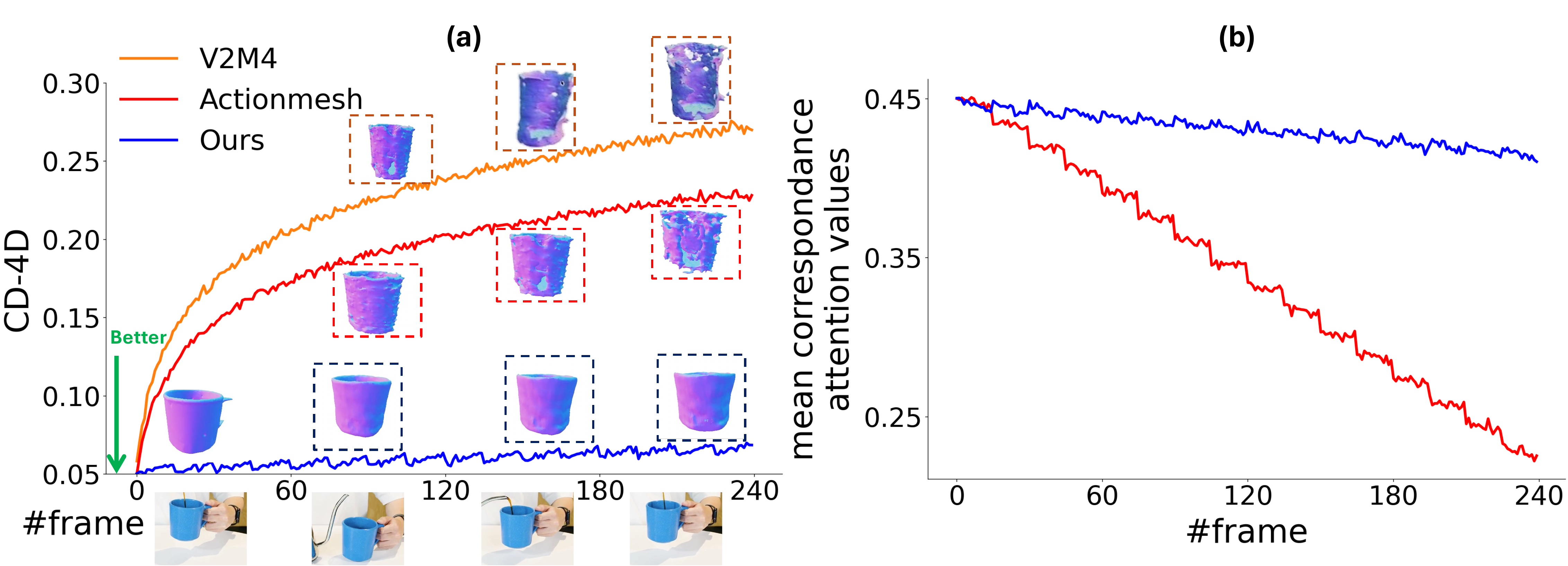}
\caption{\textbf{Long-sequence rollout.}
(a) Naive autoregressive 4D generation accumulates errors over time, degrading mesh quality. We find that this drift is driven by weakening latent correspondences across windows. (b) Our correspondence reinforcement preserves these correlations, stabilizing the rollout and maintaining high-quality generation.}
    \label{fig:long_video}
\end{figure}

Existing 4D generators are trained on short clips, so autoregressive rollout to longer videos quickly drifts: each new window is initialized from the final latent of the previous one, and errors accumulate (Fig.~\ref{fig:long_video}a). We measure this on long ActionBench~\citep{actionmesh} sequences (Appendix~\ref{sec:impl_details}) and observe both degrading mesh quality (Fig.~\ref{fig:long_video}a) and a steady drop in the correlation of matched latent points across windows (Fig.~\ref{fig:long_video}b).

To prevent this drift, we reinforce temporal correspondences during denoising inside each 16-frame window. The first two denoising steps run normally, establishing initial correspondences and confidence scores $c_v^{(f)}$. During the two remaining steps, we trace the attention paths backward to identify the main latent token pair $(t,t')$ behind each match, collect these pairs in $\mathcal{C}$, and scale the corresponding entries in $A_{Z_a\rightarrow Z_f}$ by their confidence:
\begin{equation}
\tilde{A}_{Z_a\rightarrow Z_f}[t,t'] =
\frac{c_v^{(f)} A_{Z_a\rightarrow Z_f}[t,t']}
{\sum_{k} c_v^{(f)} A_{Z_a\rightarrow Z_f}[t,k]},
\qquad \forall (t,t')\in\mathcal{C}.
\end{equation}
After these reinforced denoising steps, we use the final frame as the anchor for the next window. Boosting reliable attention paths stabilizes latent correlations and mesh quality over long sequences.

\subsection{Extension to 2D and 4D Point Tracking}\label{sec:2d_4d_tracking}
Beyond mesh animation, attention chaining provides a general composition mechanism: any two attention maps that share an intermediate representation can be linked, e.g., image patches to tokens, tokens to tokens, and tokens to vertices. We demonstrate this flexibility on two tasks: 2D point tracking in the input video, and 4D point tracking that recovers world-coordinate 3D trajectories for every visible pixel.
\paragraph{2D point tracking ($P_a \rightarrow Z_a \rightarrow Z_f \rightarrow P_f$).}We replace the 3D decoder attention with the denoiser cross-attention between latent tokens and image patches. Let $A_{P_f\rightarrow Z_f} \in \mathbb{R}^{P\times N}$ denote the attention from $P$ image patches to $N$ latent tokens in frame $f$. Given a query patch $p_a$ in the anchor frame, we transport its attention through the temporal block to find its correspondence $\tilde{p}_f$ in frame $f$:\begin{equation}
    \tilde{p}_f = \arg\max_p \sum_{t,t'=1}^{N} A_{P_a\rightarrow Z_a}^T[t, p_a]\, A_{Z_a\rightarrow Z_f}[t, t']\, A_{P_f\rightarrow Z_f}^T[t', p]
\end{equation}This directly reuses the temporal attention from Section~\ref{sec:attention_chain} to establish semantic persistence across the video sequence without any additional computation.\paragraph{2D to 3D bridge ($P \leftrightarrow V$).} By composing anchor-frame attentions, we directly link image patches and mesh vertices. Notably, ActionMesh~\citep{actionmesh} lacks this capability, leaving no direct way to map their output mesh back to the source image. In contrast, for a given anchor patch $p_a$ or vertex $v_a$, we establish the following mappings :\begin{equation}
    \tilde{v}_{p} = \arg\max_{v} \sum_{t=1}^{N} A_{P_a\rightarrow Z_a}[p_a,t]\, A_{Z_a\rightarrow V_a}[t,v], \quad \tilde{p}_{v} = \arg\max_{p} \sum_{t=1}^{N} A_{P_a\rightarrow Z_a}[p,t]\, A_{Z_a\rightarrow V_a}[t,v_a]
\end{equation}This creates a training-free correspondence layer connecting the input pixels to the canonical mesh geometry.\paragraph{Camera pose estimation.}Using this bridge, we collect 2D to 3D correspondences $\{(\mathbf{u}_v, \mathbf{V}_a[v])\}$ between anchor-frame pixel coordinates $\mathbf{u}_v$ and mesh vertices. Given camera intrinsics $K$, we estimate the camera pose $(R, t)$ via robust PnP~\citep{terzakis2020consistently,lepetit2009epnp,fischler1981ransac}:\begin{equation}
    (R^\star, t^\star) = \arg\min_{R,t} \sum_{v} \rho_v \left( \|\pi_K(R\,\mathbf{V}_a[v] + t) - \mathbf{u}_v\|_2 \right)
\end{equation}
Here $\pi_K$ is the perspective projection and $\rho_v$ is a robust weighting factor for each correspondence.
\paragraph{4D point tracking.}Finally, we lift pixels to 3D by intersecting anchor-frame rays with the anchor mesh and tracking their barycentric coordinates over time. Because the animated mesh vertices $\{\hat{\mathbf{V}}^{(f)}\}_{f=1}^F$ from Section~\ref{sec:animation} reside in a canonical object space, we apply our estimated camera pose to align them with the input video. For a pixel $\mathbf{u}$ with barycentric weights $\mathbf{w}$ on face $\phi$, its 3D trajectory in the anchor camera coordinates at frame $f$ is calculated as:\begin{equation}
    \mathbf{X}^{(f)}_{\mathbf{u}} = R^\star \sum_{i=0}^{2} w_i\, \hat{\mathbf{V}}^{(f)}_{\mathcal{F}[\phi, i]} + t^\star
\end{equation}This maps the canonical mesh deformation back into the observer's frame of reference, yielding dense 3D trajectories for all foreground pixels across the video sequence.

\section{Experiments}
\label{sec:experiments}

We evaluate our approach in three complementary settings that exercise the full attention chain introduced in Sec.~\ref{sec:method}: \textbf{(1)} \emph{4D mesh generation} (Sec.~\ref{sec:exp_animation}), \textbf{(2)} \emph{2D point tracking} on dynamic objects (Sec.~\ref{sec:exp_2d}), and \textbf{(3)} world-coordinate \emph{4D point tracking} (Sec.~\ref{sec:exp_4d}). We build on ActionMesh~\cite{actionmesh} as our base model (Sec.\ref{sec:preliminaries}) for all our experiments. For 2D and 4D  tracking where videos have more than 16 frames, we used the autoregressive scaling method presented in Sec.\ref{sec:scaling_video}. All experiments ran on an H100 GPU.

\subsection{4D Mesh Generation}
\label{sec:exp_animation}

\textbf{Datasets and metrics.} Following~\cite{actionmesh}, we evaluate on \textit{ActionBench}~\cite{actionmesh} and \textit{Consistent4D}~\cite{consistent4d}. ActionBench contains $16$-frame clips with ground-truth 4D meshes, while Consistent4D is used to evaluate out-of-distribution rendering quality. For geometry, we report end-to-end generation \textit{Time}, \textit{CD-3D} (per-frame), \textit{CD-4D} (full 4D point cloud), \textit{CD-M} (motion-only), and \textit{Normal Consistency}. All predictions are aligned to ground truth with ICP before metric computation; results without ICP are provided in the supplement. For rendering, we report \textit{LPIPS}~\cite{lpips}, \textit{CLIP}~\cite{clipscore}, and \textit{DreamSim}~\cite{dreamsim} between rendered and ground-truth views.

\textbf{Baselines.} Following~\cite{actionmesh}, we compare against state-of-the-art video-to-4D methods: Step1X-3D~\cite{step1x3d}, L4GM~\cite{l4gm}, GVFD~\cite{gvfd}, LIM~\cite{lim}, DreamMesh4D (DM4D)~\cite{dreammesh4d}, V2M4~\cite{v2m4}, ShapeGen4D (SG4D)~\cite{shapegen4d}, and ActionMesh~\cite{actionmesh}. We also include image-to-3D models applied independently per frame: TripoSG~\cite{triposg} and TRELLIS~\cite{trellis}.

\begin{table}[t]
  \centering
\caption{\textbf{Left}: ActionBench results. Our training-free approach is the fastest and achieves SoTA CD-3D, CD-4D, and Normal Consistency. \textbf{Right}: Consistent4D rendering results. With camera pose estimation (\emph{Ours + CPE}), our method achieves the best LPIPS, CLIP, and DreamSim among 4D mesh generation methods.}
  \label{tab:animation_combined}
  \begin{minipage}[t]{0.60\linewidth}
    \centering
    \resizebox{\linewidth}{!}{%
    \begin{tabular}{l c c c c c}
    \toprule
    \textbf{ActionBench} & Time (s) & CD-3D $\downarrow$ & CD-4D $\downarrow$ & CD-M $\downarrow$ & \makecell{Normal\\Consist.\,$\uparrow$} \\
    \midrule
    TRELLIS~\cite{trellis} & 900 & 0.065 & 0.181 & --    & --   \\
    TripoSG~\cite{triposg} & 120  & 0.056 & 0.184 & --    & --   \\
    \midrule
    DM4D~\cite{dreammesh4d}      & 2100 & 0.104 & 0.152 & 0.265 & --   \\
    LIM~\cite{lim}               & 900 & 0.089 & 0.126 & 0.243 & --   \\
    V2M4~\cite{v2m4}             & 2100 & 0.068 & 0.340 & 0.616 & --   \\
    SG4D~\cite{shapegen4d}       & 900 & 0.060 & 0.170 & 0.348 & 0.91 \\
    ActionMesh~\cite{actionmesh} & 120  & 0.053 & 0.081 & \textbf{0.148} & 0.85 \\
    \midrule
    \textbf{Ours} & \textbf{9} & \textbf{0.048} & \textbf{0.077} & 0.163 & \textbf{0.97} \\
    \bottomrule
    \end{tabular}}
  \end{minipage}%
  \hfill
  \begin{minipage}[t]{0.38\linewidth}
    \centering
    \resizebox{\linewidth}{!}{%
    \begin{tabular}{l c c c c}
    \toprule
    \textbf{Consistent4D} & Aligned & LPIPS\,$\downarrow$ & CLIP\,$\uparrow$ & \makecell{Dream\\Sim\,$\downarrow$} \\
    \midrule
    Step1X-3D~\cite{step1x3d}    & \xmark & 0.1524              & 0.9040 & 0.1106              \\
    L4GM~\cite{l4gm}             & \cmark & 0.0988              & 0.9397             & 0.0487              \\
    GVFD~\cite{gvfd}             & \xmark & 0.1691              & 0.8601             & 0.1467              \\
    SG4D~\cite{shapegen4d}       & \xmark & 0.1359  & 0.9009             & 0.0966              \\
    ActionMesh~\cite{actionmesh} & \xmark & 0.1458              & 0.9012             & 0.0939  \\
    \midrule
    \textbf{Ours}                & \xmark & 0.1423              & 0.9112             & 0.0823              \\
    \textbf{Ours + CPE}          & \cmark & \textbf{0.0823}     & \textbf{0.9468}    & \textbf{0.0319}     \\
    \bottomrule
    \end{tabular}}
  \end{minipage}
\end{table}

\begin{figure}[t!]
    \centering
    \includegraphics[width=1\linewidth]{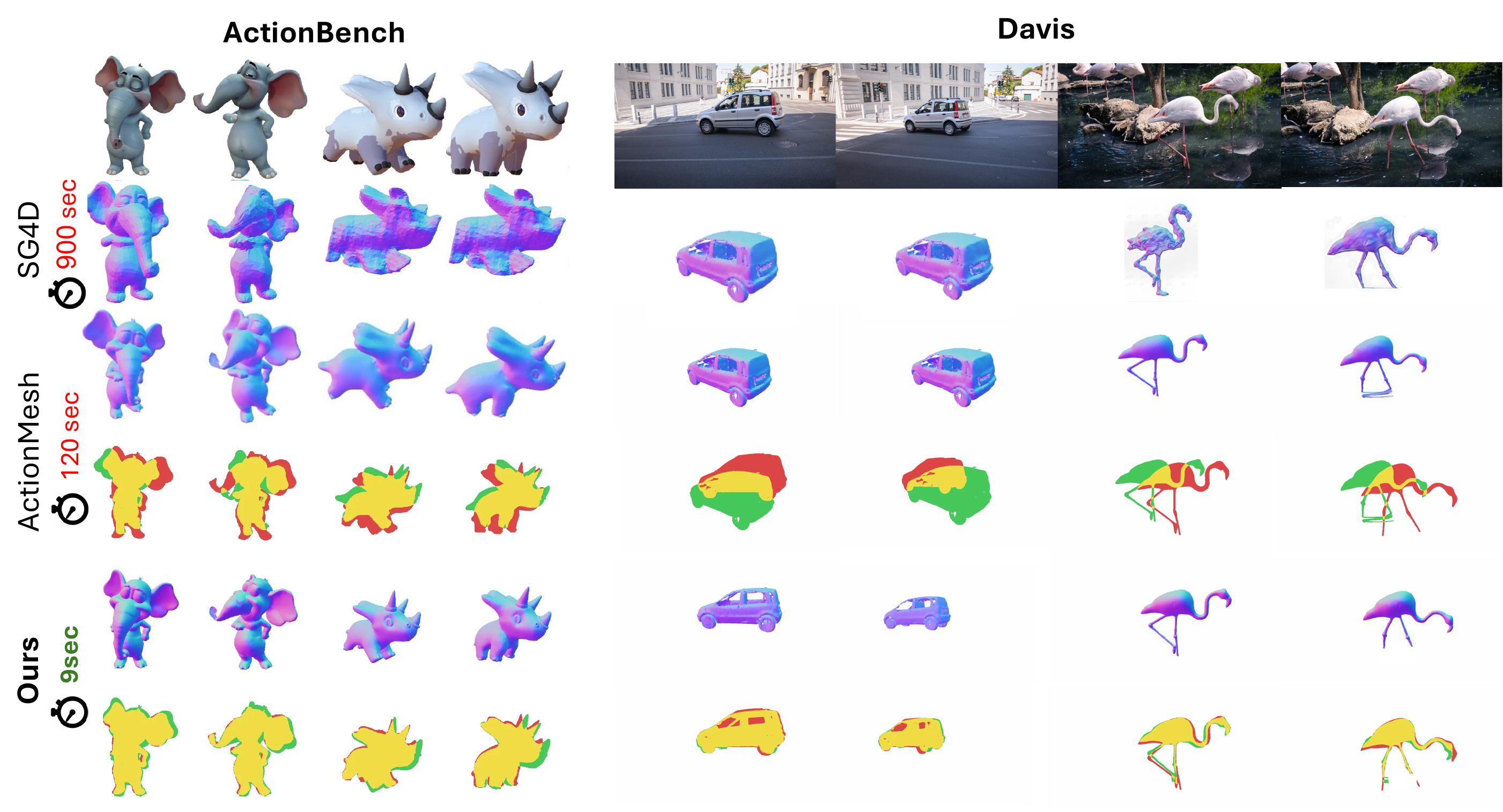}
\caption{\textbf{4D Mesh Generation.}
Our method produces sharp, temporally consistent meshes, aligns them to the input camera, and runs in \textbf{only $9$ sec}, compared to $120$--$900$ sec for prior methods. Unlike SG4D and ActionMesh, which generate object-centric meshes, our spatial attention-chain correspondences enable camera recovery and world placement, yielding high foreground overlap (yellow) and fewer mismatch regions (red/green).}
    \label{fig:qualitative_4d}
\end{figure}

\textbf{Quantitative results.}
Table~\ref{tab:animation_combined}(left) shows that our method achieves the best performance on three of four geometric ActionBench metrics: CD-3D ($0.048$), CD-4D ($0.077$), and Normal Consistency ($0.97$), while being over an order of magnitude faster: $\mathbf{9}$\,s per $16$-frame clip vs.\ $2$\,min for ActionMesh~\cite{actionmesh} ($\sim\!14\times$) and $15$\,min for ShapeGen4D~\cite{shapegen4d} ($\sim\!100\times$). ActionMesh is slightly better on CD-M ($0.148$ vs.\ $0.163$).  On Consistent4D (Table~\ref{tab:animation_combined} (right)), our method outperforms all non-aligned baselines, and with camera-pose estimation (\emph{Ours\,+\,CPE}) it surpasses the aligned baseline L4GM~\cite{l4gm} on all metrics. These results demonstrate that our method produces high-quality, topology-consistent, and camera-aligned 4D meshes, without extra training or an additional network.

\textbf{User study.} We conducted $2{,}000$ pairwise comparisons between our method and ActionMesh~\cite{actionmesh}, with $100$ raters judging appearance and motion consistency. Ours is preferred in $75\%$ of comparisons. See Supp.\ for details.

\textbf{Qualitative results.}
Fig.~\ref{fig:qualitative_4d} shows that our method produces sharp, temporally consistent meshes while also uniquely aligning them to the input camera. Unlike ActionMesh and ShapeGen4D, which generate object-centric meshes with clear silhouette mismatch, our attention-chain correspondences provide dense $2\mathrm{D}\!\leftrightarrow\!3\mathrm{D}$ matches for PnP camera recovery. This yields accurate image alignment (yellow regions) and directly enables 4D tracking and camera-aware applications.

\subsection{2D Point Tracking}
\label{sec:exp_2d}

\textbf{Datasets and metrics.} Our pipeline follows ActionMesh~\cite{actionmesh} in reconstructing a single dynamic object rather than the full scene, so we evaluate 2D tracking on foreground object points only. We use \textit{BADJA}~\cite{badja}, a standard articulated-animal joint-tracking benchmark, and a foreground-only version of \textit{TAP-Vid-DAVIS}~\cite{tapvid,davis}. Following~\cite{tapvid}, we report Average Jaccard (\textit{AJ}), average position accuracy ($\langle\delta\rangle_{\mathrm{avg}}$), and Occlusion Accuracy (\textit{OA}). On BADJA, we report segmentation accuracy on joint trajectories (\textit{segA}) and the fraction of predictions within $3$\,px of ground truth ($\delta^{3\mathrm{px}}$).

\textbf{Baselines.} We compare with three groups of trackers: \textit{2D-supervised} methods, including BootsTAP~\cite{bootstap}, TAPIR~\cite{tapir}, CoTracker~\cite{cotracker}, TAP-Net~\cite{tapvid}, PIPs~\cite{pips}, OmniMotion~\cite{omnimotion}, and CowTracker~\cite{cowtracker}; \textit{3D-aware} trackers that lift 2D tracks into 3D, including SpatialTracker~\cite{spatracker}; and \textit{zero-shot} feature-based trackers from pretrained diffusion models, DiffTrack~\cite{difftrack} and Denoise-to-Track~\cite{denoisetotrack}.

\begin{table*}[t]
  \centering
  \caption{\textbf{Point tracking results.}
  (a) and (b): 2D point tracking on DAVIS-foreground and BADJA.
  (c): 4D point tracking on PointOdyssey (PO) and Dynamic Replica (DR).}
  \label{tab:tracking_results}
  \setlength{\tabcolsep}{1pt}
  \renewcommand{\arraystretch}{0.92}

  \begin{minipage}[t]{0.36\linewidth}
    \centering
    \small
    \textbf{(a) 2D Tracking (DAVIS)}
    \begin{tabular}{l c c c}
    \toprule
     & AJ & $\langle\delta\rangle_{\mathrm{avg}}$ & OA \\
    \midrule
    \multicolumn{4}{l}{\emph{Supervised}} \\
    BootsTAP~\cite{bootstap}         & 56.64 & 69.99 & 87.88 \\
    SpatialTracker~\cite{spatracker} & 54.97 & 71.51 & 85.64 \\
    CowTracker~\cite{cowtracker}     & \textbf{60.08} & \textbf{73.53} & 88.64 \\
    \midrule
    \multicolumn{4}{l}{\emph{Zero-shot}} \\
    DiffTrack~\cite{difftrack}       & 26.44 & 43.01 & 72.10 \\
    DenoiseToTrack~\cite{denoisetotrack} & 35.15 & 50.12 & 75.16 \\
    \textbf{Ours}                    & 53.33 & 66.34 & \textbf{90.41} \\
    \bottomrule
    \end{tabular}
  \end{minipage}
  \hfill
  \begin{minipage}[t]{0.27\linewidth}
    \centering
    \small
    \textbf{(b) 2D Tracking (BADJA)}
    \begin{tabular}{l c c}
    \toprule
     & segA & $\delta^{3\mathrm{px}}$ \\
    \midrule
    \multicolumn{3}{l}{\emph{Supervised}} \\
    TAP-Net~\cite{tapvid}            & 54.4 & 6.3  \\
    PIPs~\cite{pips}                 & 61.9 & 13.5 \\
    TAPIR~\cite{tapir}               & 66.9 & 15.2 \\
    OmniMotion~\cite{omnimotion}     & 57.2 & 13.2 \\
    CoTracker~\cite{cotracker}       & 63.6 & \textbf{18.0} \\
    SpatialTracker~\cite{spatracker} & \textbf{69.2} & 17.1 \\
    \midrule
    \multicolumn{3}{l}{\emph{Zero-shot}} \\
    \textbf{Ours}                    & 64.8 & 16.3 \\
    \bottomrule
    \end{tabular}
  \end{minipage}
  \hfill
  \begin{minipage}[t]{0.29\linewidth}
    \centering
    \small
    \textbf{(c) 4D Tracking}
    \begin{tabular}{l c c}
    \toprule
     & PO & DR \\
    \midrule
    \multicolumn{3}{l}{\emph{Supervised}} \\
    St4RTrack~\cite{st4rtrack}         & 72.04 & 76.82 \\
    TraceAny.~\cite{traceanything}     & 47.02 & 61.19 \\
    Any4D~\cite{any4d}                 & 64.25 & 70.33 \\
    4RC~\cite{4rc}                     & 66.92 & 76.11 \\
    V-DPM~\cite{vdpm}                  & \textbf{82.12} & \textbf{78.18} \\
    \midrule
    \multicolumn{3}{l}{\emph{Zero-shot}} \\
    ActionMesh (II)~\cite{actionmesh}      & 31.5 & 41.6 \\
    \textbf{Ours}                      & 59.9 & 65.3 \\
    \bottomrule
    \end{tabular}
  \end{minipage}
\end{table*}

\textbf{Results.}
Table~\ref{tab:tracking_results}(a) shows that our method is the strongest zero-shot tracker on DAVIS-foreground, outperforming DiffTrack and Denoise-to-Track on all metrics. It also achieves the best OA overall ($90.41$), while remaining competitive with supervised trackers. Table~\ref{tab:tracking_results}(b), on the BADJA benchmark, our zero-shot method remains close to supervised trackers, outperforming several of them despite using no tracking supervision. These results show that attention-chain correspondences provide strong 2D tracking directly from the frozen 4D generation backbone.

\subsection{4D Point Tracking}
\label{sec:exp_4d}

\textbf{Datasets and metrics.} We follow the dynamic-only protocol of the WorldTrack benchmark~\cite{st4rtrack} and evaluate on \textit{PointOdyssey (PO)}~\cite{pointodyssey} and \textit{Dynamic Replica (DR)}~\cite{dynamicreplica}. We report \textit{APD$_{\mathbf{3D}}$}, the percentage of predictions within a set of 3D distance thresholds, averaged over thresholds after global median alignment.

\textbf{Baselines.} We compare against ActionMesh Stage~II~\cite{actionmesh} as a zero-shot baseline for per-frame mesh tracking, and five supervised 4D trackers: V-DPM~\cite{vdpm}, 4RC~\cite{4rc}, Any4D~\cite{any4d}, TraceAnything~\cite{traceanything}, and St4RTrack~\cite{st4rtrack}.

\textbf{Results.}
Tab.~\ref{tab:tracking_results}(c) shows that our method substantially improves over the zero-shot ActionMesh Stage~II baseline on both datasets (+28.4 APD$_{3D}$ on PO and +23.7 on DR). This demonstrates that attention-chain correspondences, together with camera estimation, provide a much stronger signal for world-coordinate 4D tracking.
Despite using no 4D tracking supervision, our method is competitive with supervised trackers: it outperforms TraceAnything and approaches Any4D and 4RC. Our approach close a large part of the gap without task-specific training.

\section{Conclusion}
\label{sec:conclusion}

In this paper, we introduced a training-free framework for fast video-to-4D-mesh reconstruction. By exposing spatio-temporal attention chains inside a frozen 4D backbone, we recover correspondences that animate an anchor mesh without a learned deformation network. The same mechanism yields \textit{topology-consistent meshes}, \textit{faster inference}, \textit{longer rollouts}, and enables \textit{2D/4D tracking} with \textit{camera recovery}.
At the same time, our framework inherits the limitations of the frozen models it builds on. Mesh quality depends on the image-to-3D model and denoiser; sparse smoothing and local-rigid deformation can damp fine motion; and multi-minute rollouts may still degrade as errors accumulate and attention over generated anchors becomes increasingly diffuse.

{\small
\bibliographystyle{abbrv}
\bibliography{egbib}
}

\newpage
\appendix

\begin{center}
{\LARGE\bfseries Supplementary Material}
\end{center}
\vspace{1em}

\section{Additional Qualitative Results}\label{sec:qualitative_supp}

\begin{figure}[h!]
    \centering
    \includegraphics[width=\linewidth]{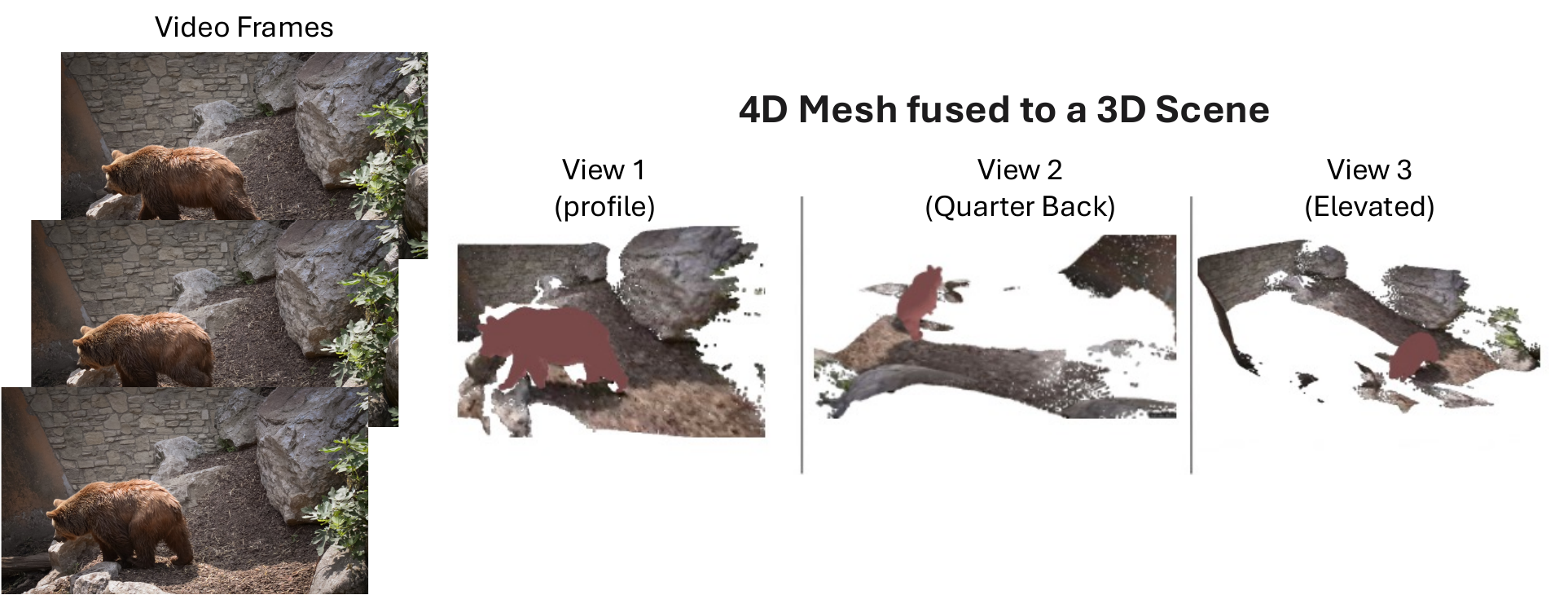}
    \caption{\textbf{Video-to-4D Scene Alignment.} Given our generated 4D object and a reconstructed background scene, we align the object to the environment using our dense 2D-to-3D correspondences. While this example utilizes Depth Anything V3~\citep{depthanything3}, our alignment framework is completely agnostic and functions seamlessly with any underlying scene reconstruction method. Please refer to our supplementary website for interactive, dynamic visualizations.}
    \label{fig:mesh_placement}
\end{figure}

We provide additional qualitative comparisons that complement the quantitative evaluation in the main paper. These results further demonstrate the advantages of our training-free attention-chain framework across four settings: 4D mesh generation, long-sequence rollout, 2D point tracking, and 4D mesh placement in reconstructed scenes.

\begin{figure}[t!]
    \centering
    \includegraphics[width=1\linewidth]{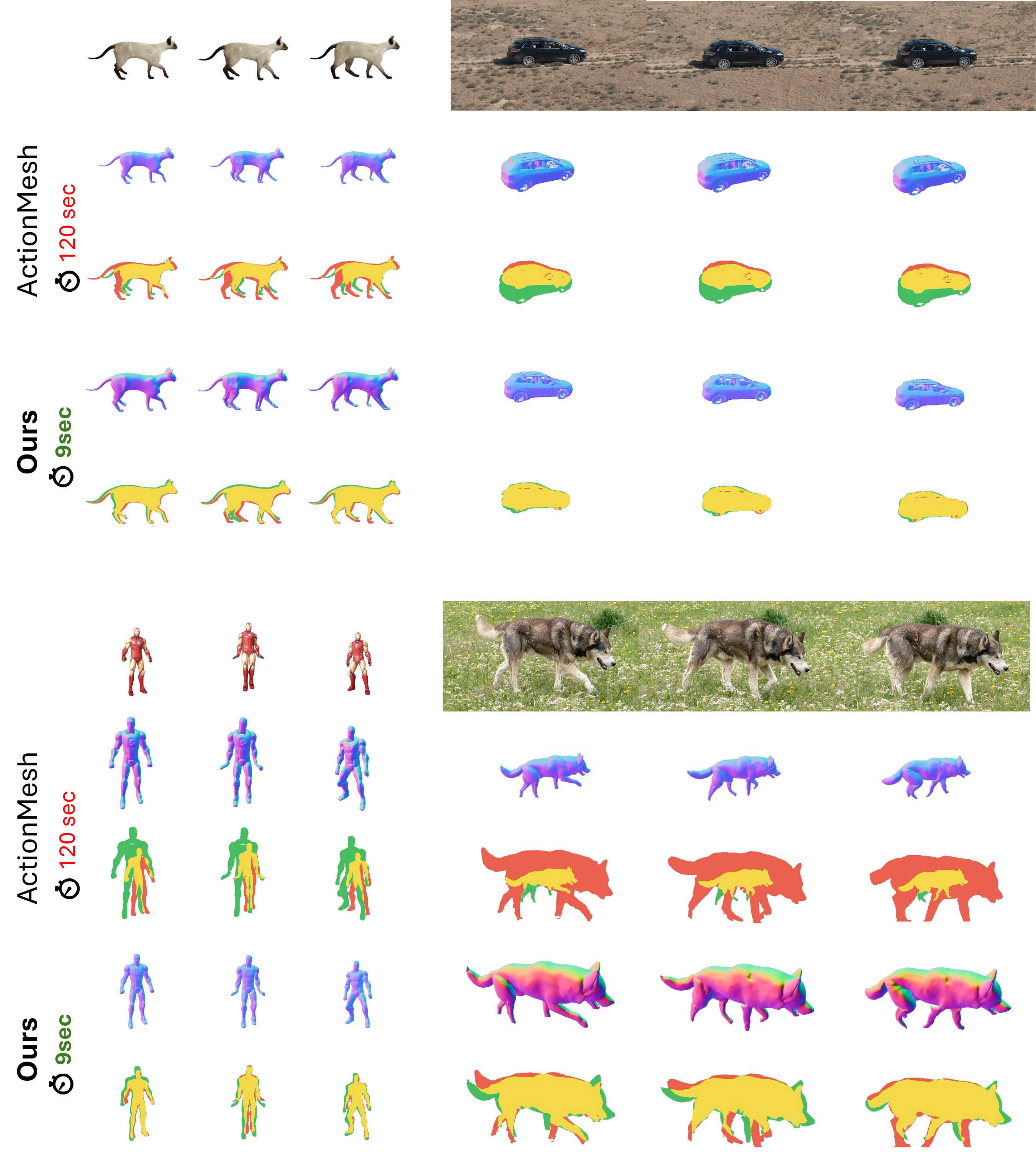}
    \caption{\textbf{Additional 4D mesh generation results.}
    Side-by-side comparison of ActionMesh~\cite{actionmesh} (left of each pair) and our method (right) on diverse ActionBench sequences. Each row shows a different input video. Our method produces sharper, temporally consistent meshes with fewer distortions while requiring only $9$\,s per clip vs.\ $120$\,s for ActionMesh.}
    \label{fig:supp_qualitative}
\end{figure}

\paragraph{4D mesh generation.}
Figure~\ref{fig:supp_qualitative} presents additional side-by-side comparisons between ActionMesh~\cite{actionmesh} and our method on sequences from ActionBench and in-the-wild videos. Our approach consistently produces comparable to better geometry with fewer temporal artifacts, while running in only $9$\,s per clip compared to ActionMesh's $120$\,s. Unlike ActionMesh, which generate object-centric meshes, our spatial attention-chain
correspondences enable camera recovery and world placement, yielding high foreground overlap
(yellow) and fewer mismatch regions (red/green)

\begin{figure}[t!]
    \centering
    \includegraphics[width=1\linewidth]{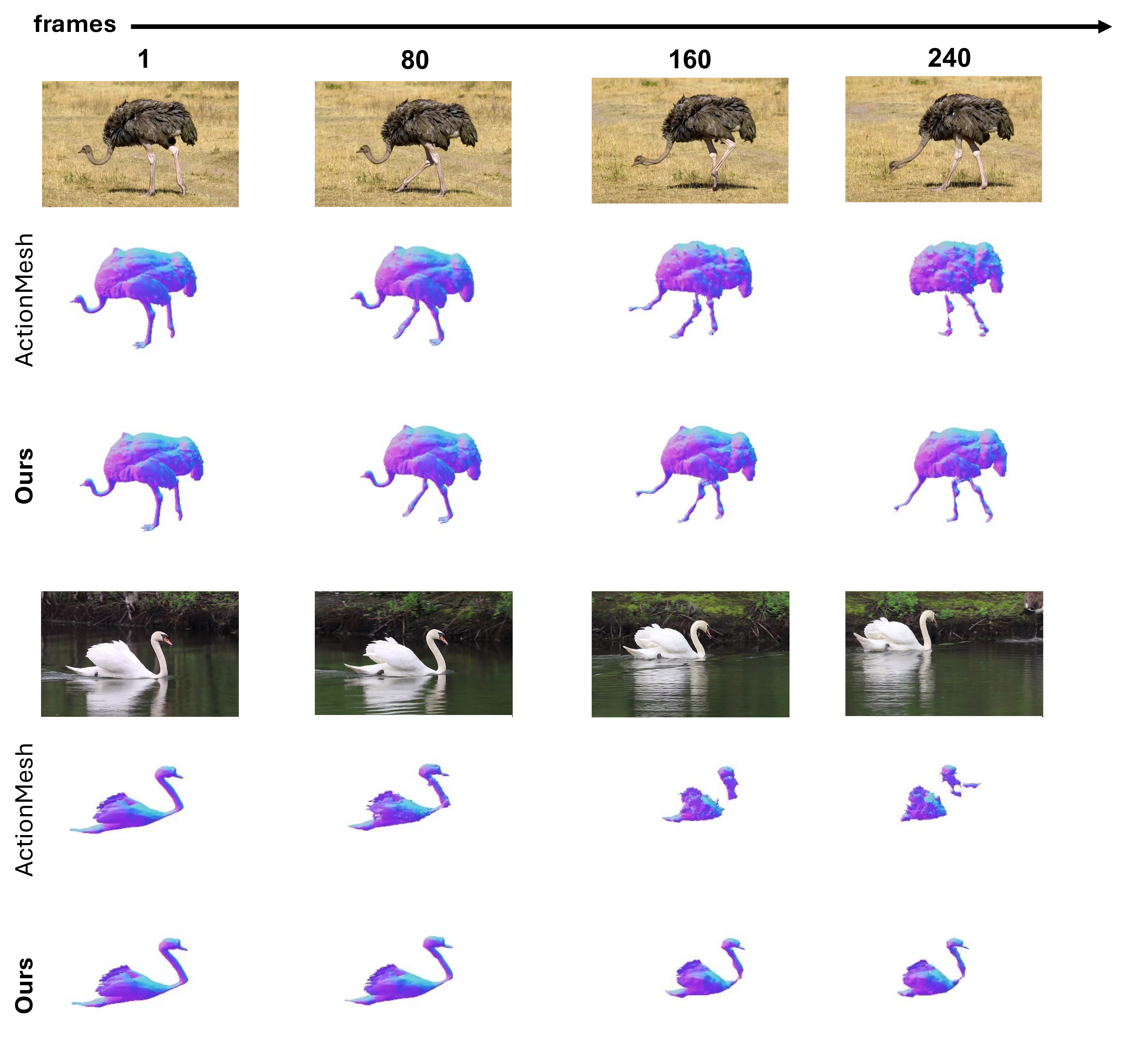}
    \caption{\textbf{Long-sequence autoregressive generation.}
    Mesh quality over $240$ frames (sampled at frames $1$, $80$, $160$, $240$). ActionMesh progressively degrades due to accumulated drift across autoregressive windows, eventually losing recognizable geometry. Our correspondence reinforcement maintains stable, high-quality meshes throughout the entire sequence.}
    \label{fig:supp_long}
\end{figure}

\paragraph{Long-sequence generation.}
Figure~\ref{fig:supp_long} shows autoregressive rollouts spanning $240$ frames. ActionMesh accumulates drift over successive windows, leading to progressive geometry degradation and loss of recognizable structure. In contrast, our correspondence reinforcement (Sec.~\ref{sec:scaling_video}) maintains stable mesh quality throughout the sequence, preserving both global pose and fine detail even at frame $240$.

\begin{figure}[t!]
    \centering
    \includegraphics[width=1\linewidth]{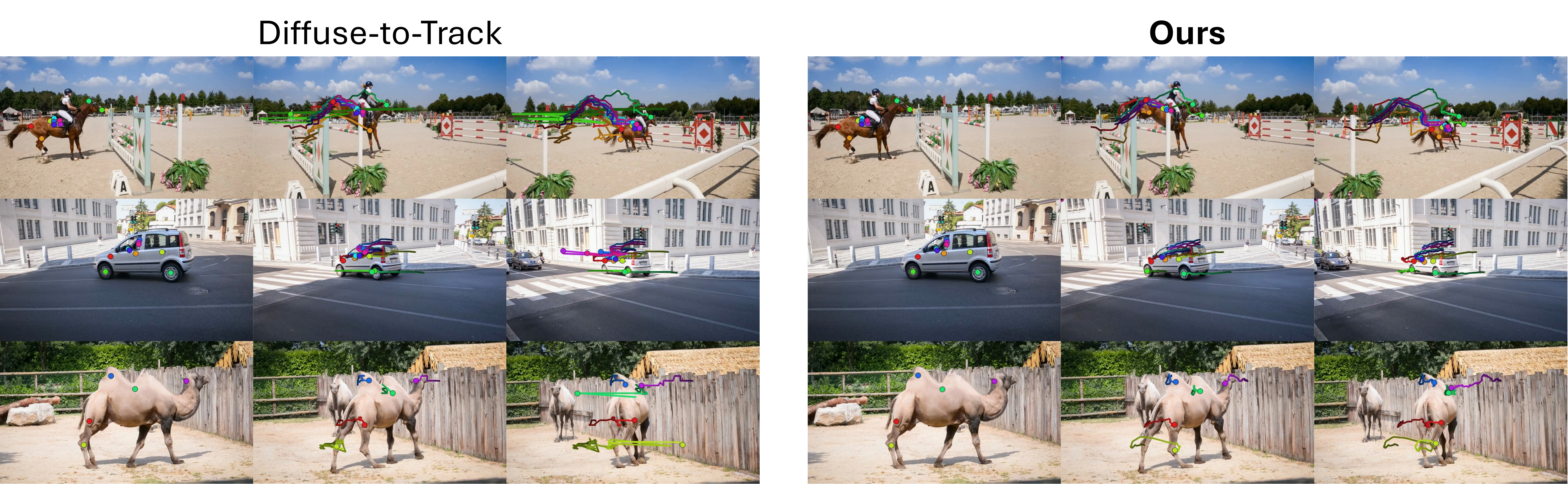}
    \caption{\textbf{Zero-shot 2D point tracking.}
    Comparison with Denoise-to-Track~\cite{denoisetotrack} on challenging sequences with articulated motion. Colored dots mark tracked query points; trails show predicted trajectories. Our attention-chain correspondences yield geometrically grounded tracks that more faithfully follow the underlying surface motion, particularly on limbs and fine structures.}
    \label{fig:supp_2d_tracking}
\end{figure}

\paragraph{2D point tracking.}
Figure~\ref{fig:supp_2d_tracking} compares our zero-shot 2D point tracking against Denoise-to-Track~\cite{denoisetotrack}, the current zero-shot state of the art. Each colored dot represents a tracked query point, with trails showing the predicted trajectory. Our method produces smoother, more accurate trajectories that better follow the underlying surface motion, particularly on articulated body parts where attention-chain correspondences provide geometrically grounded matches rather than purely appearance-based ones.

\paragraph{4D mesh placement.}
Figure~\ref{fig:mesh_placement} demonstrates that our camera pose estimation enables fusing the generated 4D mesh directly into a reconstructed 3D scene. Given only the input video frames, we recover the camera trajectory via PnP and place the animated mesh into a scene point cloud, allowing it to be viewed from arbitrary novel viewpoints. The mesh correctly occupies the physical space of the subject across all views, confirming that our attention-chain correspondences provide accurate world-space localization without any pose supervision. An interactive version with full camera control is available on the supplementary website.

\section{Ablation Study}\label{sec:ablation}

We probe three design choices: (i) the number of Stage~I denoising
steps, (ii) the resulting per-clip wall-clock vs.\ ActionMesh, and
(iii) the contribution of each component on long, $240$-frame
sequences. Unless stated otherwise we evaluate on ActionBench with the
same protocol as Section~\ref{sec:exp_animation}.

\paragraph{Denoising steps vs.\ quality.}
Both methods share the same Stage~I temporal denoiser $\Phi_\theta$;
ActionMesh further runs a learned Stage~II that regularises the
per-frame latents $Z$, while ours reads the attention of the
\emph{last} denoising step and propagates it through the chain
$\mathbf{V}\!\to\!\mathbf{T}\!\to\!\mathbf{T}\!\to\!\mathbf{V}$.
Figure~\ref{fig:metrics_steps} plots CD-3D, CD-4D and CD-Motion as a
function of the number of denoising steps. Ours plateaus by step~$4$
on every metric, whereas ActionMesh keeps improving up to $\sim
10$--$20$ steps. On CD-3D and CD-4D the gap is large in the few-step
regime that matters for inference cost: at $4$ steps ours is roughly
$2\times$ better on CD-3D ($0.048$ vs.\ $0.095$) and CD-4D ($0.077$
vs.\ $0.125$), and stays strictly better even at $30$ steps. On
CD-Motion ActionMesh is consistently slightly better -- its learned
Stage~II yields smoother frame-to-frame motion than our closed-form
correspondence chain ($0.148$ vs.\ $0.152$ at $30$ steps; $0.161$
vs.\ $0.163$ at $4$ steps), with the gap growing at very few steps
where Stage~II is most needed to clean up noisy latents. The remaining
headroom on motion smoothness is therefore in the temporal
regulariser, not in the underlying correspondence. All other
experiments use $5$ denoising steps.

\paragraph{Inference time breakdown.}
We accelerate 4D mesh generation in two complementary ways
(Table~\ref{tab:ablation_time}). \textbf{(i) Many fewer denoising
steps.} Our correspondence is robust to slightly noisy latents and
saturates by step~$4$ on every metric (Fig.~\ref{fig:metrics_steps}),
whereas ActionMesh requires $\sim$$30$ steps for its Stage~II network
to operate at peak quality -- shrinking Stage~I from $\sim$$100$\,s to
$\sim$$7.5$\,s on the same backbone. \textbf{(ii) No additional
network.} We drop ActionMesh's learned Stage~II ($\sim$$10$\,s)
entirely and replace it with a lightweight, training-free pipeline of
four closed-form operations: a single batched VAE decode for all
frames at once ($1.0$\,s), FPS landmark sampling on the anchor mesh
($0.46$\,s), the V$\!\to$T$\!\to$T$\!\to$V correspondence
computation ($0.16$\,s), and a geodesic, topology-preserving
animation ($0.8$\,s) -- altogether $\sim$$2.4$\,s. End-to-end our full
pipeline takes $\sim$$9$\,s per $16$-frame clip, vs.\ $\sim$$110$\,s
for ActionMesh: an order-of-magnitude speed-up without any loss of
quality on CD-3D / CD-4D (Sec.~\ref{sec:exp_animation}).

\textbf{Component contribution on long videos.}
Table~\ref{tab:ablation_components} ablates the three components of our
long-video pipeline on the $240$-frame split of ActionBench, evaluated
\emph{without} ICP so that the metrics reflect the \emph{intrinsic}
alignment of each variant. ActionMesh in this regime is unaligned: its
Stage~II is trained on $16$-frame clips and its autoregressive
extension drifts in absolute pose, leading to large CD and motion
errors. Replacing Stage~II with our attention-chain correspondence
(Sec.~\ref{sec:attention_chain}) tightens both metrics. Periodically
re-encoding the anchor across windows
(Sec.~\ref{sec:scaling_video}) further removes intra-AR drift. Finally,
PnP-based camera-pose estimation (Sec.~\ref{sec:2d_4d_tracking}) folds out the
remaining global rigid mismatch and gives the largest jump on CD-3D
and CD-Motion. Our full pipeline more than halves both CD and motion
error on the same backbone, \emph{without ICP}.

\section{Implementation Details}
\label{sec:impl_details}

\paragraph{Backbone and pipeline.}
We inherit the frozen, off-the-shelf two-stage backbone of
\texttt{ActionMesh}~\cite{actionmesh}: Stage~0 is
\texttt{TripoSG}~\cite{triposg}, an image-to-3D diffusion model that
produces the anchor mesh $\mathcal{M}_a$ ($\sim$$20\text{k}$ vertices
after decimation) and its latent code
$z_a \in \mathbb{R}^{N \times d}$ with $N = 2048$, $d = 64$; Stage~I
is a flow-matching temporal denoiser $\Phi_\theta$ that jointly
produces per-frame latents $Z = \{z_f\}_{f=0}^{F-1}$ in a window of
$F = 16$ frames, conditioned on per-frame
DINOv2~\cite{dinov2} patch features and anchored at $z_a$. We
\textbf{never train or fine-tune any model}: all weights are frozen,
and our entire pipeline reduces to (i)~reading the V$\!\to$T
cross-attention of the TripoSG VAE decoder and the inflated
self-attention of $\Phi_\theta$ at the first 3 denoising steps,
and (ii)~a sequence of closed-form correspondence and skinning
operations (Sec.~\ref{sec:attention_chain}). The same attention chain
serves 2D point tracking, 4D point tracking, and camera-pose
estimation (Sec.~\ref{sec:2d_4d_tracking}). For 2D and 4D tracking under occlusions or unstable object masks, we first apply object an amodal completion using video diffusion model~\cite{taco} before running our correspondence
and deformation pipeline. For sequences longer than
$F$ frames we run a sliding $16$-frame window with periodic anchor
re-encoding (Sec.~\ref{sec:scaling_video}).

\begin{table}[t]
\centering
\caption{Per-clip wall-clock breakdown ($16$ frames). Ours accelerates
generation in two ways: (i)~Stage~I with $4$ denoising steps instead
of $30$; (ii)~replacement of ActionMesh's learned Stage~II network
with a lightweight, training-free pipeline of four cheap closed-form
operations. The two together reduce per-clip latency from
$\sim$$110$\,s to $\sim$$9$\,s.}
\label{tab:ablation_time}
\setlength{\tabcolsep}{8pt}
\renewcommand{\arraystretch}{1.10}
\small
\begin{tabular}{l c c}
\toprule
                                                       & ActionMesh~\cite{actionmesh} & Ours \\
\midrule
\textbf{Stage~I} (per-frame latents $Z$)               & $100.00$ \emph{(30 steps)}   & $\;\;7.50$ \emph{(4 steps)} \\
\midrule
\textbf{Stage~II} (per-frame meshes)                   & $\;\;15.00$ \emph{(learned)} & $\;\;1.49$ \emph{(training-free)} \\
\quad batched VAE decode (all frames at once)          & --                           & $\;\;0.87$                       \\
\quad FPS landmarks (anchor mesh)                      & --                           & $\;\;0.46$                       \\
\quad correspondence computation (attention chain)                       & --                           & $\;\;0.16$                       \\
\quad geodesic topology-preserving animation           & --                           & $\;\;0.005$                       \\
\midrule
\textbf{Total}                                         & $110.00 s$                     & $\;\;\mathbf{9.0 s}$              \\
\bottomrule
\end{tabular}
\end{table}

\begin{table}[t]
\centering
\caption{Component ablation on $240$-frame sequences, \emph{no ICP}.
We start from ActionMesh's unaligned predictions and add our
components one-by-one. Lower is better.}
\label{tab:ablation_components}
\setlength{\tabcolsep}{8pt}
\small
\begin{tabular}{l c c c}
\toprule
Configuration & CD-3D $\downarrow$ & CD-4D $\downarrow$ & CD-M $\downarrow$ \\
\midrule
ActionMesh~\cite{actionmesh} (unaligned)                       & 0.260 & 0.260 & 0.373 \\
\quad + temporal correspondence (Sec.~\ref{sec:attention_chain}) & 0.190 & 0.195 & 0.310 \\
\quad + long-video AR (Sec.~\ref{sec:scaling_video})                & 0.155 & 0.162 & 0.250 \\
\quad + camera-pose estimation (Sec.~\ref{sec:2d_4d_tracking})             & \textbf{0.108} & \textbf{0.115} & \textbf{0.198} \\
\bottomrule
\end{tabular}
\end{table}

\begin{figure}[t]
  \centering
  \includegraphics[width=\linewidth]{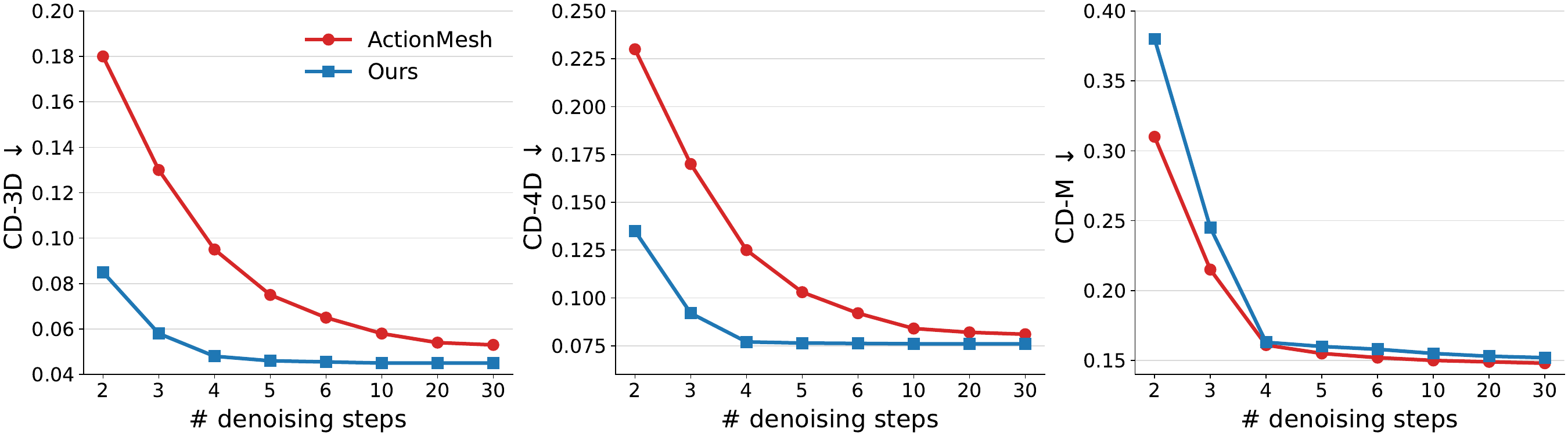}
  \caption{CD-3D / CD-4D / CD-Motion vs.\ Stage~I denoising steps on
  ActionBench. Ours plateaus by step~$4$ and decisively wins CD-3D and
  CD-4D at every step count; ActionMesh's learned Stage~II is
  consistently smoother on CD-Motion.}
  \label{fig:metrics_steps}
\end{figure}

\paragraph{Hyperparameters.}
After a coarse grid search we
fixed a single configuration that we use across \emph{every}
experiment in this paper -- 4D mesh generation, 2D and 4D point
tracking, and the long-video stress tests -- with no per-scene
tuning. The values are: $4$ Stage~I
denoising steps (justified in Fig.~\ref{fig:metrics_steps}); $1000$ FPS landmarks on the anchor mesh with
curvature boost $\alpha = 2.0$; outlier rejection at
$5\times$ the per-landmark median displacement and a $\geq\!95\%$
valid-frame requirement; Gaussian smoothing of landmark displacements
with $\sigma_{\text{landmark}} = 1.5$ and of the full vertex
displacement field with $\sigma_{\text{final}} = 1.0$;
a geodesic-rigid skinning with $k_{\mathrm{NN}} = 120$ Dijkstra
neighbours (auto $\sigma$).

For the camera intrinsics, we assume a fixed focal length of 2.1875 defined in canonical Normalized Device Coordinates (NDC). Because the NDC image extent spans [-1, 1], this value is independent of pixel resolution and corresponds to a full vertical field-of-view of roughly 49.2°. Camera poses are then estimated using a robust PnP solver implemented as a RANSAC loop with EPnP as the minimal solver, an 8-pixel reprojection-error inlier threshold, $400$ iterations at $0.999$ confidence, an SQPnP consistent-solver fallback when EPnP returns a degenerate (near-camera-at-infinity) translation, and a final Levenberg--Marquardt refinement on the inlier set.

\paragraph{Synthetic Long Sequences}
In Section~\ref{sec:scaling_video}, we generate extended ActionBench~\citep{actionmesh} sequences by employing a \emph{ping-pong} looping scheme. Because standard ActionBench sequences are limited to exactly 16 frames, we artificially lengthen the temporal duration of the videos by continuously playing the frames forward and then in reverse. Specifically, the sequence ordering follows the alternating pattern $(f_0, f_1, \dots, f_{14}, f_{15}, f_{14}, \dots, f_1, f_0, f_1, \dots)$. This progression allows us to generate arbitrarily long sequences while preserving smooth and continuous motion transitions.

\section{Topology-Preserving Animation Details}\label{sec:appendix_animation}

This section provides the mathematical details for the temporal smoothing and mesh deformation steps introduced in Section~\ref{sec:animation}.

\subsection{1D Gaussian Temporal Smoothing}

The raw trajectories of the control landmarks, extracted from the attention chain, are denoted by $\tilde{\mathbf{v}}^{(f)}_\ell$ for landmark $\ell$ at frame $f$. To reject physically implausible predictions, we apply an outlier filtering step. Specifically, we flag a landmark's displacement as an outlier if its magnitude exceeds a relative threshold (e.g., $5\times$ the median displacement magnitude of all landmarks in that frame) or an absolute distance threshold. We assign a confidence score $c_\ell^{(f)}$ to each prediction, where $c_\ell^{(f)} = 0$ if the displacement is flagged as an outlier, and $c_\ell^{(f)} = 1$ otherwise.

To bridge the gaps caused by rejected outliers and ensure fluid motion, we apply a confidence-weighted 1D Gaussian temporal smoothing independently to each landmark's trajectory. Crucially, we smooth the \textit{displacements} from the anchor pose rather than the absolute positions. This ensures that stationary landmarks do not artificially drift over time.

The smoothed position $\hat{\mathbf{v}}^{(f)}_\ell$ for landmark $\ell$ at frame $f$ is computed as:
$$
\hat{\mathbf{v}}^{(f)}_\ell
=
\mathbf{v}^a_\ell
+
\frac{
\sum_j G_\sigma(f,j)\,c_\ell^{(j)}
\left(\tilde{\mathbf{v}}^{(j)}_\ell-\mathbf{v}^a_\ell\right)
}{
\sum_j G_\sigma(f,j)\,c_\ell^{(j)}+\epsilon
},
$$
where $\mathbf{v}^a_\ell$ is the position of the landmark in the anchor mesh, $G_\sigma(f,j) = \exp\left(-\frac{(f - j)^2}{2\sigma^2}\right)$ is a 1D Gaussian kernel with standard deviation $\sigma$ (in frames), and $\epsilon$ is a small constant for numerical stability. The anchor frame itself is pinned to its original geometry to prevent any smoothing leakage.

\subsection{Local Rigid Deformation}

To propagate the smoothed landmark motions to the dense mesh, we use a local rigid deformation formulation related to As-Rigid-As-Possible (ARAP) surface modeling~\cite{arap}. Standard linear blend skinning interpolates displacement vectors, which famously causes volume loss when regions undergo rotation. By instead solving for a local rigid transformation (rotation and translation) for each vertex, we preserve the local volume of the mesh.

For each free vertex $v$ on the anchor mesh, we identify its $K_{nn}$ geodesically closest landmarks, denoted as the neighborhood $\mathcal{N}_v$. We assign a Gaussian weight to each landmark $\ell \in \mathcal{N}_v$ based on its geodesic distance $d_{v\ell}$ on the anchor mesh:
$$
w_{v\ell} = \exp\left(-\frac{d_{v\ell}^2}{2\sigma_{geo}^2}\right),
$$
where $\sigma_{geo}$ is a scaling factor proportional to the mean spacing between landmarks. The weights are normalized such that $\sum_{\ell \in \mathcal{N}_v} w_{v\ell} = 1$. Using geodesic distances ensures that the deformation respects the surface topology and articulations (e.g., preventing a landmark on the torso from inappropriately influencing the arm).

For each frame $f$, we solve a weighted Procrustes alignment to find the optimal rotation $R_v^{(f)} \in SO(3)$ that maps the anchor landmark positions to their smoothed target positions:
$$
R_v^{(f)}
=
\arg\min_{R\in SO(3)}
\sum_{\ell\in\mathcal{N}_v}
w_{v\ell}
\left\|
R(\mathbf{v}^a_\ell-\boldsymbol{\mu}^a_v)
-
(\hat{\mathbf{v}}^{(f)}_\ell-\boldsymbol{\mu}^{(f)}_v)
\right\|_2^2,
$$
where $\boldsymbol{\mu}^a_v$ and $\boldsymbol{\mu}^{(f)}_v$ are the weighted centroids of the neighborhood in the anchor and target frames, respectively:
$$
\boldsymbol{\mu}^a_v
=
\sum_{\ell\in\mathcal{N}_v}
w_{v\ell}\mathbf{v}^a_\ell,
\qquad
\boldsymbol{\mu}^{(f)}_v
=
\sum_{\ell\in\mathcal{N}_v}
w_{v\ell}\hat{\mathbf{v}}^{(f)}_\ell.
$$

Once the optimal rotation $R_v^{(f)}$ is found via Singular Value Decomposition (SVD), we apply the rigid transformation to the vertex's anchor position to obtain its final animated position:
$$
\hat{\mathbf{v}}^{(f)}_v
=
R_v^{(f)}(\mathbf{v}^a_v-\boldsymbol{\mu}^a_v)
+
\boldsymbol{\mu}^{(f)}_v.
$$
Applying this transformation to all vertices yields the final animated mesh $\hat{\mathcal{M}}_f = (\hat{\mathbf{V}}^{(f)}, \mathcal{F}_a)$.

\subsection{User Study Details}
\label{sec:user_study}

To complement the geometric metrics reported in Sec.~\ref{sec:exp_animation}, we conducted a perceptual user study that directly compares our 4D mesh generation against ActionMesh~\cite{actionmesh}, the strongest baseline in this setting.  We randomly sampled $20$ clips for the study and for each clip, both methods were run with their default settings on the same input frames; the resulting 4D meshes were rendered from the input camera and assembled into a single side-by-side video that places the two renderings next to the reference input. Each trial shows this triplet to the rater, with the left/right ordering of the two methods randomised per trial to remove positional bias, and asks: \emph{``Which result better matches the reference video in terms of appearance and motion consistency (i.e., fewer temporal mesh distortions)?''} The rater must pick exactly one of the two options. We recruited $100$ independent raters, each judging all $20$ clips, which yields $2{,}000$ pairwise comparisons in total. Across this entire pool, $85\%$ of the judgments favour our method, and the preference is consistent across clips and categories. This perceptual margin mirrors the quantitative findings of Tab.~\ref{tab:animation_combined}: our renderings exhibit sharper geometry, more accurate silhouette alignment with the input view, and noticeably fewer temporal jitters and mesh distortions than ActionMesh.

\end{document}